\documentclass[conference]{IEEEtran}
\IEEEoverridecommandlockouts
\usepackage{cite}
\usepackage{amsmath,amssymb,amsfonts}
\usepackage{algorithmic}
\usepackage{graphicx}
\usepackage{textcomp}
\usepackage{xcolor}
\usepackage{url}
\usepackage{booktabs}
\usepackage{multirow}
\usepackage{float}

\def\BibTeX{{\rm B\kern-.05em{\sc i\kern-.025em b}\kern-.08em
    T\kern-.1667em\lower.7ex\hbox{E}\kern-.125emX}}
\begin{document}

\title{NAP3D: NeRF Assisted 3D-3D Pose Alignment for Autonomous Vehicles\ \

}


\author{
\IEEEauthorblockN{Gaurav Bansal}
\IEEEauthorblockA{
\textit{MIT THINK Scholars Program}\textsuperscript{\dag}, Archbishop Mitty High School \\
San Jose, California, USA \\
bansal22.gaurav@gmail.com
}
}

\maketitle

\renewcommand{\thefootnote}{\dag}
\footnotetext{This work was supported by the MIT THINK collegiate research program, which provided funding that enabled the project to be completed independently.}
\renewcommand{\thefootnote}{\arabic{footnote}}  

\maketitle

\begin{abstract}
Accurate localization is essential for autonomous vehicles, yet sensor noise and drift over time can lead to significant pose estimation errors, particularly in long-horizon environments. A common strategy for correcting accumulated error is visual loop closure in SLAM, which adjusts the pose graph when the agent revisits previously mapped locations. These techniques typically rely on identifying visual mappings between the current view and previously observed scenes and often require fusing data from multiple sensors.

In contrast, this work introduces NeRF-Assisted 3D-3D Pose Alignment (NAP3D), a complementary approach that leverages 3D-3D correspondences between the agent's current depth image and a pre-trained Neural Radiance Field (NeRF). By directly aligning 3D points from the observed scene with synthesized points from the NeRF, NAP3D refines the estimated pose even from novel viewpoints, without relying on revisiting previously observed locations.

This robust 3D-3D formulation provides advantages over conventional 2D-3D localization methods while remaining comparable in accuracy and applicability. Experiments demonstrate that NAP3D achieves camera pose correction within 5 cm on a custom dataset, robustly outperforming a 2D-3D Perspective-N-Point baseline. On TUM RGB-D, NAP3D consistently improves 3D alignment RMSE by approximately 6 cm compared to this baseline given varying noise, despite PnP achieving lower raw rotation and translation parameter error in some regimes, highlighting NAP3D's improved geometric consistency in 3D space. By providing a lightweight, dataset-agnostic tool, NAP3D complements existing SLAM and localization pipelines when traditional loop closure is unavailable.

\end{abstract}

\begin{IEEEkeywords}
NeRFs, calibration, SLAM, loop closure, view synthesis
\end{IEEEkeywords}
\section{Introduction}

\subsection{Motivation}
In recent times, developments in autonomous vehicle technologies have accelerated. Such systems lie in UAVs, autonomous drones, recent breakthroughs with driverless cars with Waymo and Tesla, high school robotics competitions, and many more. As robot autonomy grows, accurate navigation and localization become increasingly significant subproblems. A fundamental problem that the control systems that power autonomous vehicles try to solve is the issue of positional accuracy. One popular strategy is to localize to pre-existing mappings of traversed environments, of which Neural Radiance Fields (NeRFs) \cite{mildenhall2020nerfrepresentingscenesneural}  act as compelling options.

\subsection{Problem}
Maintaining a low loss and error has been a strong topic of research in the past; an inability to refine position in an autonomous agent can lead to catastrophic results. Additionally, sensors used to gather data for the agent's odometry, such as accelerometers or wheel encoders, may be noisy, leading to inaccurate results. Noise filters such as the Kalman Filter \cite{kalman1960} exist, but in emergency situations, an agent would benefit from being able to follow trajectories effectively without using sensors that may pose a liability.

\subsection{Contributions}
This paper outlines a method of geometry-reliant pose correction, building atop of NeRFs for localization to a reconstructed scene. Our method, NAP3D, aims to reduce the error between the agent's odometry and the actual position of the agent through image processing on camera input and a NeRF synthesized view. NeRFs, which are neural networks from which 3D views of an environment can be created, effectively act as a way to anchor the agent in its environment. By comparing the image generated by the agent at its true position to the image generated in the Neural Radiance Field at its estimated position, the transforms that dictate translation and rotation between the two images, and thus the positional error, can be effectively calculated. This comparison is done through Procrustes Analysis on corresponding 3D keypoints between the NeRF and real-life images. This introduces a more direct geometric alignment between the observed scene and the NeRF, relying on 3D-3D correspondences between keypoints rather than traditional 2D-3D projections; this allows the system to more robustly estimate translational and rotational offsets, particularly in novel viewpoints or texture-poor regions where 2D-3D methods can fail. This corrective framework only relies on a depth camera as a necessary sensor, and remains comparable to existing methods. Additionally, NAP3D can be further extended to other radiance field technologies such as Gaussian Splatting \cite{kerbl20233dgaussiansplattingrealtime}, as it only requires the fundamental properties of depth, opacity, and RGB data.

\section{Background}

\begin{figure*}
        \centering
        \centerline{\fbox{\includegraphics[width=1\textwidth]{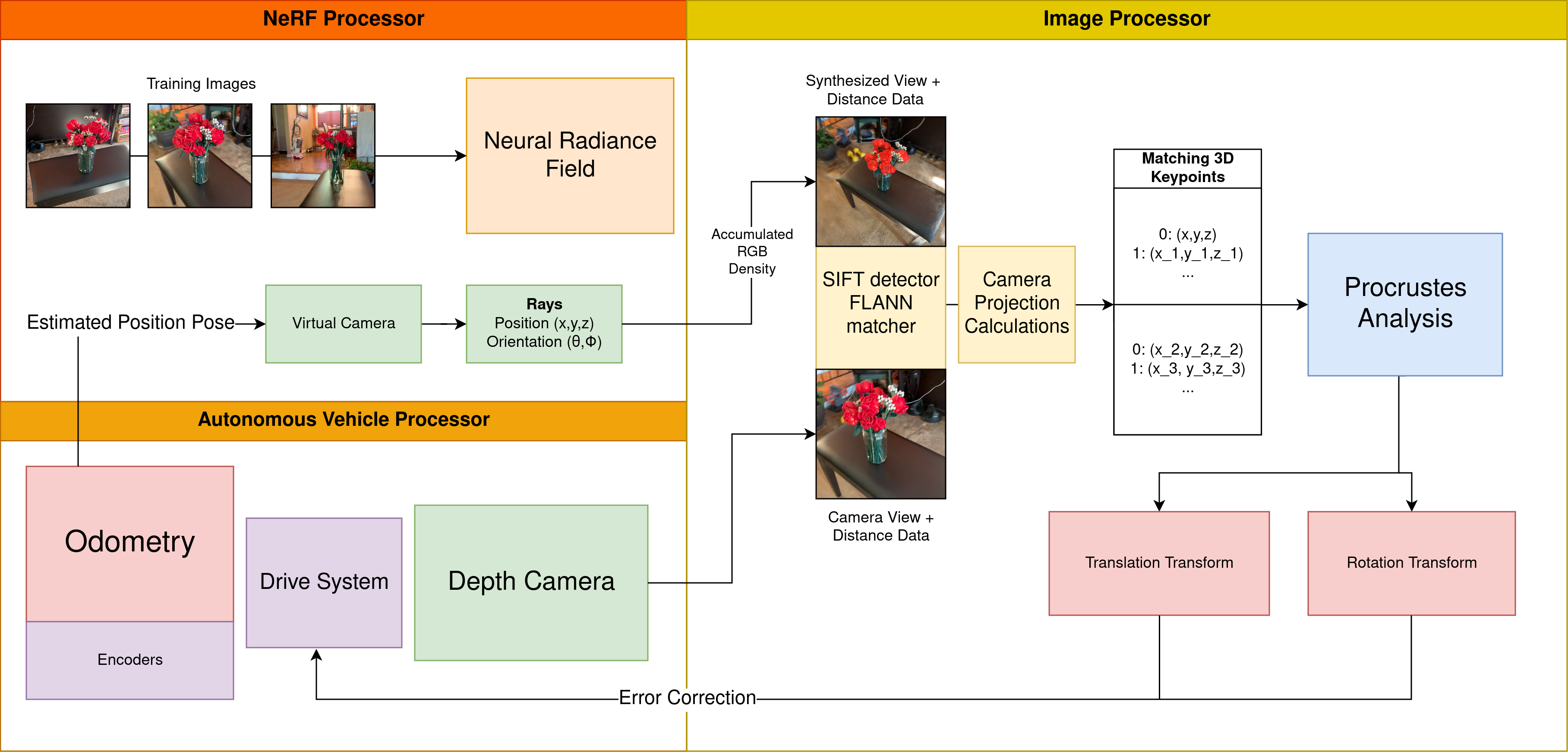}}}
        \caption{Diagram illustrating outlined algorithm structure. System is structured into the Neural Radiance Field, Onboard Image, and Autonomous Agent processors, each of which control separate parts of the system's logic. }
        \label{fig:enter-label}
\end{figure*}

\subsection{Neural Radiance Fields}   
Neural Radiance Fields have emerged as an efficient method for compact 3D scene representation and novel view synthesis. By training on multiple 2D images, NeRFs can generate new perspectives of a scene, even from viewpoints not included in the input data. Each scene is modeled as a continuous function queried along rays cast from camera pixels, taking in 3D position $(x, y, z)$ and viewing direction $(\theta, \phi)$ to output color $(R, G, B)$ and density $\sigma$. These predictions are produced by a multi-layer perceptron (MLP) and rendered using volumetric techniques, with surface depth inferred from accumulated density along rays.

NeRFs offer lightweight, high-fidelity synthesis compared to alternatives like photogrammetry, especially when training data is limited \cite{rs16020301, li2024nerfxlscalingnerfsmultiple}. For this reason, NeRFs are well-suited to applications like autonomous vehicles and form the foundation of the correctional system described here. This work uses Nerfstudio \cite{Tancik_2023}, a PyTorch-based \cite{NEURIPS2019_9015} framework with a user-friendly GUI and API for training, rendering, and programmatic NeRF generation.

\subsection{Prior Work}
Systems such as Visual-Inertial Odometry (VIO) and sensor fusion techniques like the Extended Kalman Filter (EKF) aim to refine an agent's pose estimate during motion by mitigating sensor noise and drift. These approaches are tightly integrated into the odometry computation itself, inherently relying on continuous sensor feedback. By contrast, the proposed system operates as an external correction mechanism, designed to periodically realign the agent's pose with its actual position by retroactively correcting accumulated drift atop an otherwise faulty or imprecise odometry pipeline. Hence, this design is more conceptually aligned with loop closure \cite{cummins2008fab} and global localization strategies.

Loop closure in SLAM detects when an agent revisits a location and updates the pose graph. Visual loop closure matches keypoints, often using Bag-of-Words (BoW) \cite{GalvezTRO2012} to encode descriptors for nearest-neighbor retrieval.

Recent work on Neural Radiance Fields introduces new opportunities for spatial understanding and relocalization. NeRFs can reconstruct photorealistic 3D environments from monocular image sequences and have proven to be useful in SLAM systems \cite{rosinol2022nerfslamrealtimedensemonocular}. Furthermore, loop closure in adjacent NeRF-driven systems has been researched; such systems have been improved on through the addition of semantic guidance mechanisms to reduce false matches in perceptually ambiguous areas \cite{10935649}.

Our work is functionally similar to methods such as SplatLoc \cite{zhai2024splatloc3dgaussiansplattingbased}, which also utilizes a radiance field for localization; however, while SplatLoc's methods utilize 2D-3D point correspondence, our method is different in that we leverage full 3D-3D comparison for robust geometric comparison. 

\section{Algorithm Design}
\begin{figure*}
        \centering
        \centerline{\fbox{\includegraphics[width=0.8\textwidth]{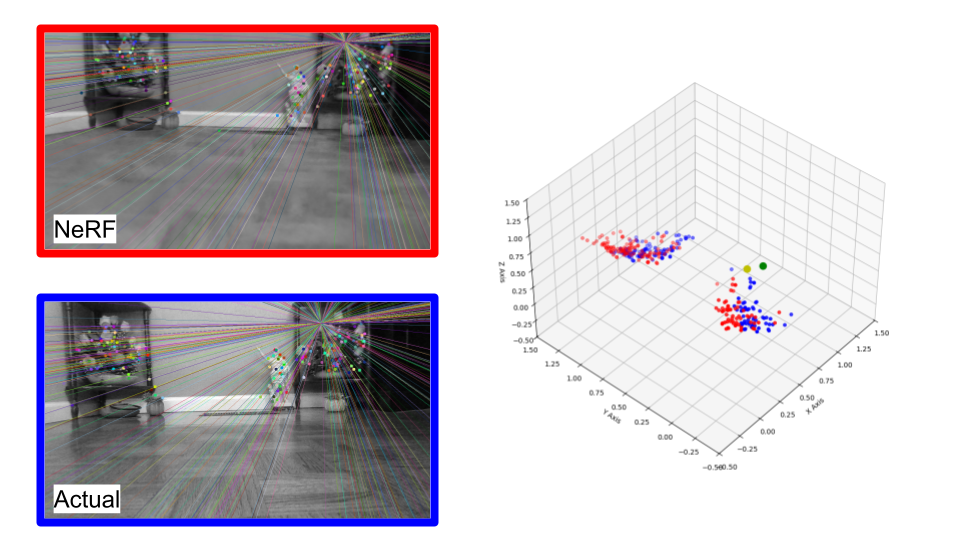}}}
        \caption{Images generated from the NeRF and RealSense, with corresponding keypoints indicating the epipole of each image. Point clouds indicating 3d coordinates of NeRF (red) and real-life (blue) keypoints are shown, with marked centroids for NeRF and RealSense in green and yellow respectively. }
        \label{fig:enter-label}
    \end{figure*}
\subsection{Algorithm Overview}
The proposed system in theory consists of three primary sections -- one to handle the Neural Radiance Field, one to perform image processing, and one to control the physical autonomous agent. Figure 1 shows a diagrammed overview of the system. The Neural Radiance Field is initially trained on an array of images and camera poses of the traversal environment. The outlined system then generates an image at the agent's estimated position through the NeRF, and calculates a depth map; similarly, an image and depth map is generated at the agent's true position through the usage of a depth camera. Image processing is performed and keypoints are generated to correlate the two images; after depth value conversion to a standard unit , Procrustes analysis yields optimal rotation, scale, and translation matrices that map one image to another, and this data is applied to the autonomous agents' movement system to correct accumulated error.

\subsection{Neural Radiance Field Setup}
The system uses Nerfstudio for NeRF training, specifically the depth-nerfacto method based on DS-NeRF \cite{deng2024depthsupervisednerffewerviews}, which emphasizes depth supervision during training. This is well suited for 3D-3D alignment, as accurate depth is critical. Integration with Zoe depth estimation \cite{bhat2023zoedepthzeroshottransfercombining} further improves depth estimation over baseline NeRFs.

The system employs Nerfstudio with the depth-nerfacto method, based on DS-NeRF \cite{deng2024depthsupervisednerffewerviews}, which emphasizes depth supervision and is well suited for 3D-3D alignment tasks where accurate depth is essential. Since no depth maps were supplied in our custom dataset, the ZOE depth estimation method \cite{bhat2023zoedepthzeroshottransfercombining} was used by depth-nerfacto to supplement training. This setup represents a lower-bound, worst-case scenario, evaluating system performance when only RGB images and camera poses are provided. We assess the full capability of depth-nerfacto in assisting NAP3D through our additional evaluation with the TUM RGB-D dataset \cite{sturm12iros}.

To gather images for our custom dataset, the mobile application KIRI Engine was used. This was due to KIRI Engine's ability to export raw camera poses in addition to frames, paired with Nerfstudio's ability to process KIRI Engine data relatively quickly.

\subsection{Neural Radiance Field Integration}
In order to generate a view from a certain position within the scene using Neural Radiance Fields, a camera must be generated at that position with supplied intrinsic parameters.

Using the Nerfstudio API, a camera could be created programatically. This camera's view rays would then be processed through the NeRF, returning accumulated outputs such as RGB data, depth, and opacity. Furthermore, \texttt{expected-depth} is used rather than \texttt{depth} in the selected output due to the higher accuracy achieved. This is mainly because while the latter acts as the distance until accumulated density reaches a sharp threshold (which is vulnerable to noise or artifacts) the former is more robust as output is smoothed and effectively denoised through averaging.

An image and depth map data can thus be created using the Nerfstudio API, and can be run through the image processing system.

\subsection{Depth Camera Integration}
For evaluation on our custom dataset, a depth camera was used to generate images and depth maps at the agent's true position in the environment. The selected depth camera was an Intel RealSense D455i \cite{keselman2017intelRealSensestereoscopicdepth} due to its accuracy at depths from 0.6-2 meters and integrated IMU. The RealSense development utilities also proved to be useful, whether it be getting camera intrinsic and extrinsic values or generating images programatically.

Because the depth camera and RGB camera present on the Intel RealSense D455i have different intrinsic parameters, and there exists an extrinsic transform between them, frame alignment must be performed. Frame alignment is handled via the Intel RealSense SDK's built-in alignment utilities.\footnote{\url{https://github.com/IntelRealSense/libRealSense}} Frame alignment allows for depth values to be accurately procured given coordinates to a pixel in an RGB image.

\subsection{Image Processing and Keypoint Generation}

Relevant keypoints between the NeRF image and real life image are generated through the Scale Invariant Feature Transform (SIFT) algorithm \cite{lowe1999object}, in combination with the Fast Library for Approximate Nearest Neighbours (FLANN) \cite{muja_flann_2009}. SIFT involves subtracting Gaussian blurred versions of the input image to find possible keypoints and distinct features. We deliberately use SIFT in place of newer learning-driven alternatives such as SuperPoint \cite{detone2018superpointselfsupervisedpointdetection} to isolate geometric effects from learned correspondence biases, and highlight NAP3D's performance when operating solely on geometric information. The FLANN algorithm matches features between each image given a threshold of accuracy, producing a set of corresponding keypoints from the NeRF-generated image to the physically generated image. Figure 2 shows NeRF-synthesized and real-life images of a scene with matching keypoints and epilines.

Given 2D keypoint coordinates, corresponding depth values, and camera intrinsics for both the NeRF and RealSense, respective 3D coordinates in camera space are calculated as per classical pinhole camera projection equations. This is all performed after depth values in both the NeRF and RealSense model are normalized to meters, either through an experimental regression or unit conversion.

The 3D coordinate $(X,Y,Z)$ in camera space is computed given 2D pixel coordinates $(x,y)$ and depth $d$ using camera intrinsic parameters:

\begin{equation}
X = \frac{(x-c_x)*d}{f_x},\quad Y=\frac{(y-c_y)*d}{f_y},\quad Z=d
\end{equation}

where $(c_x, c_y)$ are the principal point offsets and $(f_x, f_y)$ are the focal lengths.

\subsection{Procrustes Transformation and SVD}
To align the 3D keypoints derived from the NeRF and RealSense images and calculate the rotational and translational transforms, rigid SE(3) Procrustes analysis is performed. This method estimates the optimal translation between the two 3D coordinate maps that minimizes the Frobenius norm of the residuals, and transitively, the mean squared distance between aligned keypoints.

Specifically, given NeRF coordinates $\{X\}$ and RealSense coordinates $\{Y\}$, the aim is to solve the rigid Procrustes problem:

\begin{equation}
\min_{R, t} \sum_i \left\| R x_i + t - y_i \right\|^2
\end{equation}
\[
\text{subject to} \quad
R^\top R = I, \quad \det(R) = 1
\]

Which yields rotation $R \in SO(3)$ and translation $t \in \mathbf{\rm I\!R}^3$ that minimize the Frobenius norm of the residual and thus ensure accurate alignment between NeRF and RealSense coordinates.

Rigid Procrustes analysis, following the classical Umeyama method \cite{88573} of point alignment, calculates translation by comparing centroids, and rotation by performing Singular Value Decomposition (SVD) of the covariance matrix (between centered point sets).

The solution proceeds as follows, given two point sets $X = [x_1, \ldots, x_N], Y = [y_1, \ldots, y_N] \in \mathbf{\rm I\!R}^{3\times{N}}$ and \( \mathbf{1} \) as a vector of ones of length \( N \):

\begin{enumerate}
    \item Compute the centroids of each point set:
    \begin{equation}
    \bar{x} = \frac{1}{N} \sum_{i=1}^N x_i, \quad
    \bar{y} = \frac{1}{N} \sum_{i=1}^N y_i.
    \end{equation}
    \item Center the point sets by subtracting their centroids:
    \begin{equation}
    \tilde{X} = X - \bar{x} \mathbf{1}^\top, \quad
    \tilde{Y} = Y - \bar{y} \mathbf{1}^\top,
    \end{equation}
    \item Compute the covariance matrix:
    \begin{equation}
    H = \tilde{X} \tilde{Y}^\top.
    \end{equation}
    \item Perform singular value decomposition (SVD) of the covariance matrix:
    \begin{equation}
    H = U \Sigma V^\top.
    \end{equation}
    \item Compute the optimal rotation matrix:
    \begin{equation}
    R = V U^\top.
    \end{equation}
    If \( \det(R) < 0 \), correct for improper rotation by flipping the sign of the last column of \( V \) before computing \( R \).
    \item Compute the translation vector:
    \begin{equation}
    t = \bar{y} - R \bar{x}.
    \end{equation}
\end{enumerate}

This yields the rigid transformation \( (R, t) \) that best aligns the two point sets.

Figure 2 shows NeRF and real-world point clouds from our custom dataset, prior to transforms being applied. The translation and rotation between the two point sets represent the accumulated pose error between the agent's true and estimated positions. These values can enable manual pose correction in real-world navigation.

In a second evaluation on the TUM RGB-D sequence \texttt{freiburg3\_long\_office\_household}, we augment the closed-form Procrustes alignment with a RANSAC-based outlier rejection scheme \cite{10.1145/358669.358692} to improve robustness under realistic sensing noise. Rather than assuming uniformly reliable correspondences, this method repeatedly estimates the SE(3) alignment from minimal 3D--3D subsets and selects the consensus transformation that maximizes inlier support under an anisotropic 3D residual model. For a candidate transformation $(R, t)$, each correspondence $(\mathbf{x}_i, \mathbf{y}_i)$ is evaluated via a normalized residual

\begin{equation}
    r_i = \sqrt{\frac{\left\lVert (Rx_i + t - y_i)_{xy}\right\rVert^2}{\sigma_{xy}^2} + \frac{(Rx_i + t - y_i)^2_z}{\sigma_z^2}},
\end{equation}

which accounts for differing noise characteristics in lateral and depth directions. Inliers are selected based on this anisotropic confidence metric rather than a fixed isotropic threshold, and the final pose is refined using a weighted Procrustes estimate over the consensus set, where correspondence weights are derived from NeRF opacity values. This modification requires no changes to the underlying alignment model, yet significantly improves stability in the presence of depth noise, imperfect correspondences, and NeRF reconstruction artifacts, enabling consistent pose refinement.

\subsection{Complexity and Runtime}

We next analyze the computational cost of the proposed alignment stage, separate from feature extraction, correspondence matching, and NeRF rendering, which are orthogonal to the core contribution. 

Given a set of $N$ established 3D-3D correspondences, a single Procrustes alignment requires $O(N)$ time, as the singular value decomposition is performed on a fixed $3\times3$ matrix. When included in a RANSAC procedure with $K$ iterations, the total alignment cost scales as $O(KN)$, with $K$ fixed across all experiments. 

As mentioned earlier, this analysis excludes the cost of correspondence generation (e.g., keypoint extraction and matching) and NeRF-assisted generation, which are required by baseline localization approaches. In practice, the alignment stage executes efficiently on a CPU, making NAP3D suitable for online pose correction.

\section{Experimental Implementation}

\subsection{Implementation Overview}

The following evaluation includes, as alluded to earlier, two implementations regarding testing. 

The first involves a custom dataset, constructed in a closed dining room. Physical testing was performed to gather images, depth maps from the RealSense, and measure physical distances. We will refer to this as evaluation J.

To isolate software performance from hardware-induced variables, all testing with this first dataset was conducted in a controlled environment without full deployment on a physical agent. This setup enabled focused evaluation of visual input and transformation accuracy without interference from actuation or sensor noise.

NeRFs were generated using the Nerfstudio Python API with images and poses captured via the KIRI Engine app, using the depth-nerfacto pipeline. RGB and depth data were processed using Nerfstudio and NumPy \cite{harris2020array}. True-positiomn data was gathered through use of a RealSense depth camera in the room itself.

For the second evaluation, we utilized the TUM RGB-D SLAM dataset, specifically that of \texttt{freiburg3\_long\_office\_household}. This specific environment was chosen due to provision of ground truth camera positions, pre-alignment of depth maps and RGB images, and a feature-rich environment. We will refer to this as evaluation K.

A NeRF was generated using Nerfstudio on every other frame of this dataset. Depth maps supplied by TUM RGB-D were utilized by depth-nerfacto to assist robust scene depth reconstruction. NeRF-generated depth values and RGB images were compared to frames already provided by the dataset.

For both evaluations, keypoints were extracted and matched using OpenCV's SIFT/FLANN tools \cite{opencv_library}. Procrustes alignment and the baseline Pnp was implemented manually in NumPy and OpenCV, and Matplotlib \cite{Hunter:2007} was used for visualization. In evaluation K, both NAP3D and the baseline involved RANSAC outlier rejection and iteration.

\subsection{Unit Conversion}

As our method utilizes 3D-3D point correspondence, a method of converting NerF units to meters was required for accurate comparison. 

For evaluation J, we leveraged classical linear regression \cite{55e7ba22-38fb-3d2b-9a2c-0e68080abfc3} to convert NeRF depth data to depth camera data (which was in meters). The two values were gathered at 16 points in the testing scene, yielding the following regression of the form
\begin{equation}
 Y = aX + b
\end{equation}
where $ a = 1.13584$, $b = 0.0203827$, $X$ represents NeRF depth values and $Y$ represents meter values.

The analysis yielded a statistically significant relationship with an $R^2$ of $0.789$. 

For evaluation K, we evaluated comparisons in the context of the \texttt{scale} factor provided by Nerfstudio's dataparser. As the TUM depth values were converted into meters during our preprocessing step, this value was $\sim0.3718$ NeRF units per meter.

\subsection{Experimental Structure - Evaluation J}
Two testing sublocations were selected to evaluate system performance across varying distances and visual feature densities, illustrating the method's applicability across different scene setups. This setup provides a clear, dataset-agnostic baseline within a single, representative environment. The sublocations are outlined as follows:
\begin{enumerate}
  \item Sublocation A: sublocation A includes two flower pots, among various other distinct objects such as a teapot and statue. Testing was performed at closer range than sublocation B.
  \item Sublocation B: sublocation B includes a single flower pot, among some of the previously mentioned miscellaneous objects. Testing was performed at a farther range than sublocation A.
\end{enumerate}

To provide a comparative baseline, we also computed camera poses using standard 2D-3D correspondences with the OpenCV solvePnP (Perspective-n-Point) algorithm, using the same 2D keypoints projected into the RealSense images with 3D coordinates from the NeRF. RMSE was calculated for these baseline poses to quantify localization error relative to our 3D-3D alignment method. This offers a simple point of comparison to conventional 2D-3D methods on a custom dataset without implying that any specific prior work employs the same procedure.

The NeRF virtual camera was aligned with a real-world position, and trials involved placing the Intel RealSense camera at known offsets. Output translational errors were compared against these known values to assess accuracy.

Due to flat terrain, analysis focused on the horizontal $x$ and $z$ axes, omitting $y$. The system was validated through root-mean-square-error (RMSE) calculations.

\subsection{Experimental Structure - Evaluation K}

As mentioned earlier, evaluation K utilized the TUM RGB-D \texttt{freiburg3\_long\_office\_household} dataset. RGB, depth, and ground truth values were aligned by timestamp through the script provided by TUM. This yielded $>2000$ frames; for ease of training on constrained hardware, every other frame was ignored when constructing a custom \texttt{transforms.json} for Nerfstudio.

\begin{figure}
    \centering
    \includegraphics[width=1\linewidth]{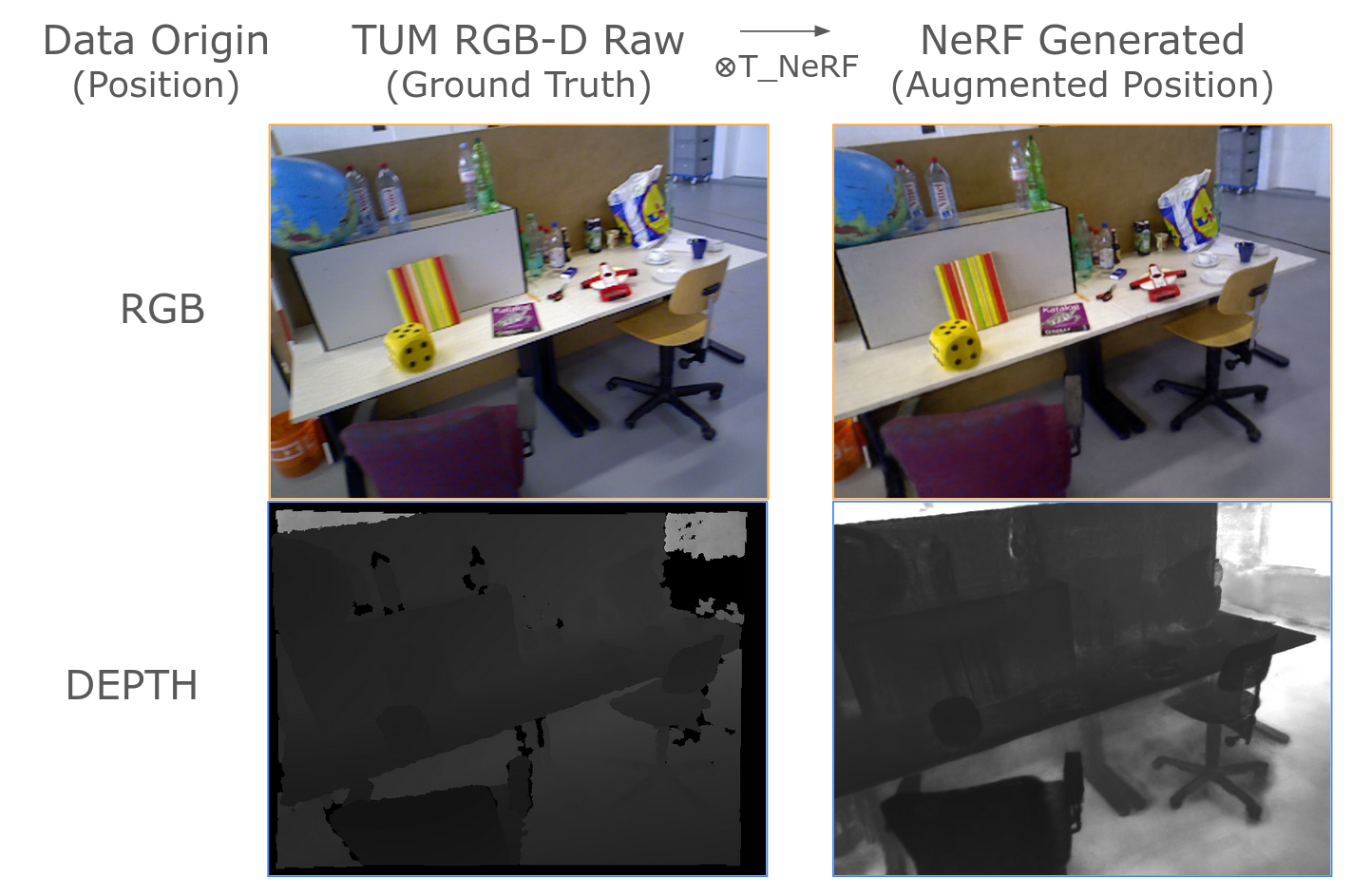}
    \caption{TUM RGB-D and NeRF RGB and Depth images at the ground truth position and perturbed position respectively. Note that while both depth images represent data fidelity, the TUM RGB-D depth image is formatted as to encode actual depth values in a 16-bit grayscale format.}
    \label{fig:placeholder}
\end{figure}
As mentioned earlier, Procrustes Analysis was reinforced through a RANSAC-driven outlier rejection scheme. To provide a comparative baseline, we performed 2D-3D comparison through OpenCV's RANSAC assisted solvePnP algorithm. This baseline utilized depth from the Neural Radiance Field to generate 3D points. 

A script was written to, given the list of frames and ground truth poses in this transforms file, generate 40 different scenarios. For each scenario, a frame was randomly chosen, and data gathering involved:
\begin{enumerate}
    \item Collecting the RGB image at ground truth provided by TUM
    \item Collecting Depth values at ground truth provided by TUM (direct processing of the 16-bit grayscale image allowed for easier serialization)
    \item Generating a random perturbation (translation in the range of $-10$ to $10$ cm; rotation in the range of $-5^\circ$ to $5^\circ$) $T_{gt}$, and converting it to NeRF units through dataparser transforms to get $T_{NeRF}$
    \item Collecting the NeRF-generated RGB image at the position resulting from augmenting the ground truth with $T_{NeRF}$
    \item Collecting the NeRF-generated depth values at this same position
    \item Collecting NeRF opacity (\texttt{accumulation}) at this same position
    \item Serializing the ground truth and $T_{gt}$ value for future error calculations
\end{enumerate}

\begin{figure*}
    \centering
    \centerline{\fbox{\includegraphics[width=1\textwidth]{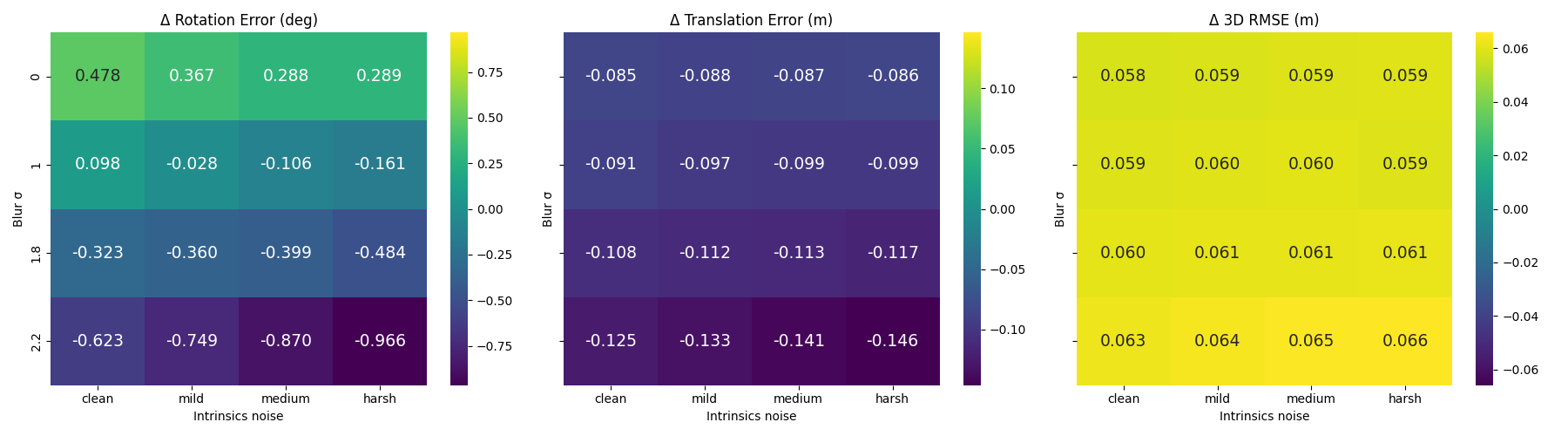}}}
    \caption{Mean per-frame difference between PnP-RANSAC and NAP3D error (PnP - NAP3D). Positive values per cell imply smaller NAP3D error; Negative values imply larger.}
    \label{fig:placeholder}
\end{figure*}

In an effort to evaluate each method's robustness with varying levels of fidelity and noise, each scenario evaluation was performed $16$ times; with a combination of $4$ image blur factors and $4$ camera intrinsics shifts. These changes were meant to simulate defocus, improper generation, and drift that could lead to incorrect output. Blur was performed through simple Gaussian Blur, with $\sigma \in \{0.0, 1.0, 1.8, 2.2\}$. Camera intrinsics were shifted with focal point scale and principal point offset $f,c \in \{(1.0, 0.0), (1.03, 2.0), (1.06, 5.0), (1.10, 10.0)\}$, forming \texttt{CLEAN}, \texttt{MILD}, \texttt{MEDIUM}, and \texttt{HARSH} intrinsics.

For each scenario, both NAP3D and PnP-RANSAC were employed to calculate pose deviation. Rotational error, translational error, and 3D RMSE were reported.

\section{Data Analysis - Evaluation J}
\subsection{Benchmarks}

\begin{table}[h]
\centering
\begin{tabular}{cccccc}
\toprule
\textbf{Trial} & \textbf{True (X,Z)} & \textbf{Est. (X,Z)} & \textbf{$\Delta$(X,Z)} & \textbf{RMSE} & \\
\midrule
A1 & (-0.120, 0.360) & (-0.192, 0.349) & (0.072, 0.011) & \multirow{3}{*}{0.046} & \\
A2 & (0.000, 0.000) & (0.012, 0.000) & (0.012, 0.000) & \\
A3 & (-0.120, 0.000) & (-0.091, 0.004) & (0.029, 0.004) & \\
\midrule
B1 & (0.100, 0.200) & (0.090, 0.182) & (0.010, 0.018) & \multirow{3}{*}{0.018} & \\ 
B2 & (-0.010,-0.480) & (-0.012, -0.493) & (0.002, 0.013) & \\
B3 & (0.195, -0.430) & (0.189, -0.449) & (0.006, 0.019) & \\
\bottomrule
\end{tabular}
\vspace{0.5em}
\caption{Per-trial error for estimated vs. true $(x, z)$ deviations in our 3D-3D correction system. All units are in meters.}
\label{tab:rmse-table}
\end{table}

\begin{table}[h]
\centering
\begin{tabular}{cccccc}
\toprule
\textbf{Trial} & \textbf{True (X,Z)} & \textbf{Est. (X,Z)} & \textbf{$\Delta$(X,Z)} & \textbf{RMSE} & \\
\midrule
A1 & (-0.120, 0.360) & (-0.130, 0.325) & (0.010, 0.035) & \multirow{3}{*}{0.049} & \\
A2 & (0.000, 0.000) & (0.006, -0.015) & (0.006, 0.015) & \\
A3 & (-0.120, 0.000) & (-0.047, -0.015) & (0.073, 0.015) & \\
\midrule
B1 & (0.100, 0.200) & (0.098, 0.204) & (0.002, 0.004) & \multirow{3}{*}{0.147} & \\ 
B2 & (-0.010,-0.480) & (-0.153, -0.377) & (0.143, 0.103) & \\
B3 & (0.195, -0.430) & (0.016, -0.391) & (0.179, 0.039) & \\
\bottomrule
\end{tabular}
\vspace{0.5em}
\caption{Per-trial error for estimated vs. true $(x, z)$ deviations in the PnP-based 2D-3D correction system. All units are in meters.}
\label{tab:rmse-table}
\end{table}
In testing, as shown in Tables I and II, NAP3D exhibited a location-wide RMSE value of $\sim0.046$ meters for sublocation A and $\sim0.018$ meters for sublocation B. The baseline 2D-3D system exhibited an RMSE value of $\sim0.049$ meters for sublocation A and $\sim0.147$ meters for sublocation B.

\subsection{Analysis}

For our 3D-3D correction system, both sublocations exhibited RMSE below $5$ cm, indicating effective error correction in real-world conditions. In comparison, the 2D-3D PnP baseline showed RMSE of approximately $\sim5$ cm and $\sim14.7$ cm for sublocations A and B, respectively, highlighting the improved accuracy of the 3D-3D alignment approach. The reduction in error is particularly pronounced in sublocation B, where fewer keypoints were available, suggesting that 3D-3D correspondences are more robust to sparser feature distributions.

In sublocation A, the higher number of keypoints did not correspond to lower error; RMSE averaged approximately $4.6$ cm. This is likely due to innate RealSense depth restrictions and the closer imaging distances to dense, feature-rich objects, which increased sensitivity to depth noise and the frequency of minor inconsistencies in 3D reconstruction. In sublocation B, observations from slightly larger distances resulted in lower RMSE of roughly $1.8$ cm for our system, while the baseline remained higher. These results suggest that 3D-3D alignment can provide more stable retroactive pose correction across varying ranges and feature densities compared with conventional 2D-3D methods.

\section{Data Analysis - Evaluation K}
\subsection{Benchmarks}

As shown in Figure 4, NAP3D maintained a consistent $6$-$7$cm decrease in 3D RMSE in comparison to PnP-RANSAC on all scenarios of TUM RGB-D. Furthermore, NAP3D outperforms PnP-RANSAC on relative rotational error given less noisy scenarios. All of this occurs despite having $\sim8$ to $14$cm more translation error, and at most $\sim1^\circ$ of rotational error.

\subsection{Analysis}

\begin{figure}
    \centering
    \includegraphics[width=1\linewidth]{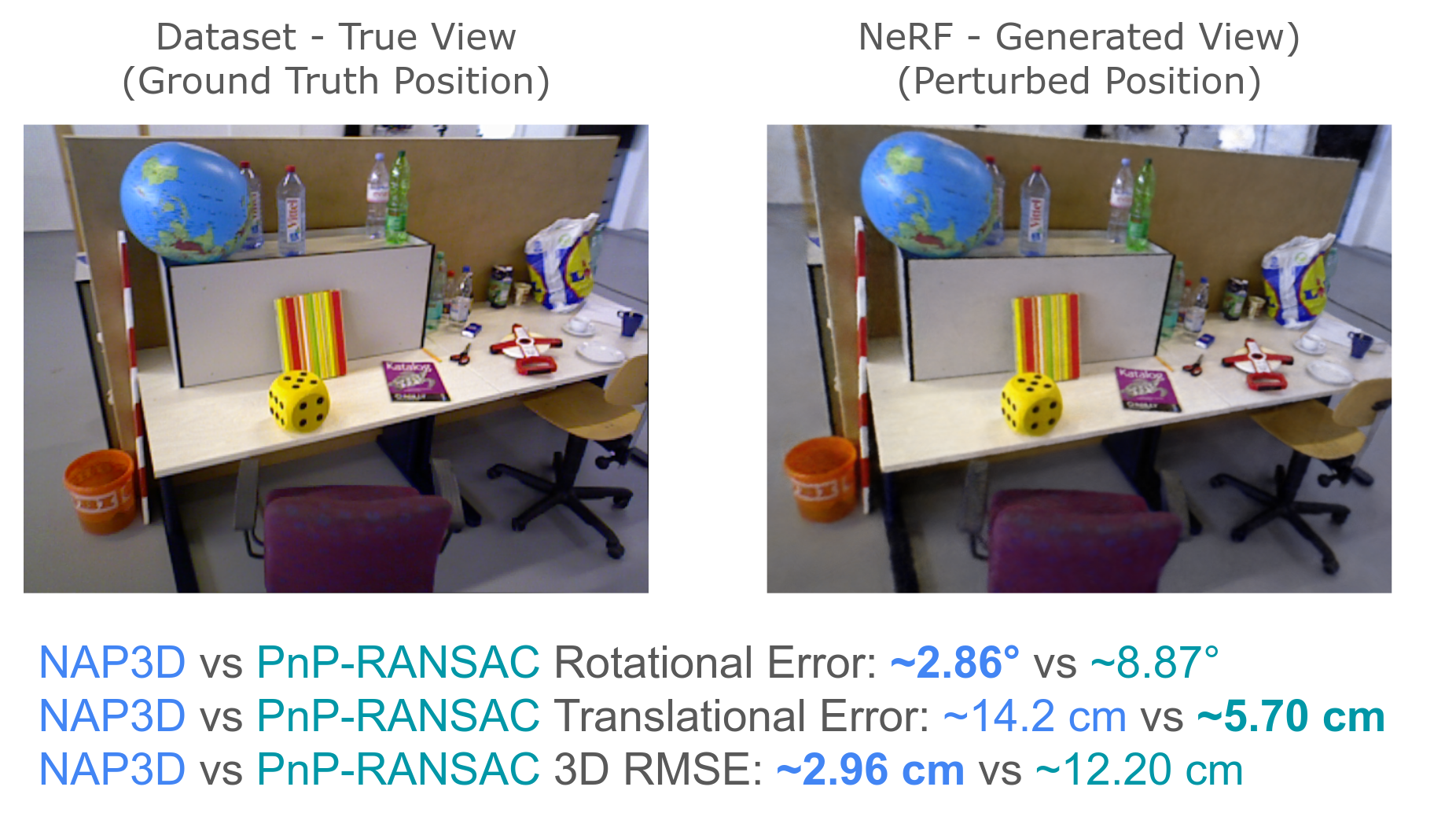}
    \caption{Qualitative comparison of pose correction, in a frame with no blur or intrinsics noise applied. PnP-RANSAC's substantially higher rotational error leads to pronounced geometric misalignment and increased 3D RMSE, although this method achieves lower translational error. In contrast, NAP3D preserves rotational consistency and achieves significantly lower 3D alignment error (decrease by $\sim9.24$ cm), illustrating a failure mode of 2D-3D pose estimation that is mitigated by 3D-3D alignment.}
    \label{fig:placeholder}
\end{figure}

On a standard dataset, NAP3D both surpasses and falls short of PnP-RANSAC depending on the evaluation metric. Positive values of 3D RMSE differences in the range of approximately $5.8$-$6.6$ cm indicate more consistent 3D alignment compared to PnP-RANSAC. Although the baseline achieves marginally lower rotational and translational error in some regimes, it consistently exhibits higher 3D RMSE.

We hypothesize that this behavior is largely attributable to higher rotational error by PnP-RANSAC in comparison to NAP3D, which induces amplified spatial deviations in projected 3D points, especially as depth increases. As shown in Figure 5, significantly larger rotational error presented by PnP-RANSAC in comparison to NAP3D ($\sim8.87^\circ$ vs. $\sim2.86^\circ$) results in substantially higher 3D RMSE despite lower translational error. This highlights the sensitivity of 3D alignment error due to angular misestimation, where even modest rotational inaccuracies can dominate spatial error in reconstructed geometry, and illustrates a regime in which NAP3D's 3D-3D formulation offers improved robustness relative to a 2D-3D PnP-RANSAC baseline. Furthermore, this difference in 3D RMSE can be credited to extreme pose estimates produced by PnP-RANSAC, whereas NAP3D produces more frequent but comparatively smaller errors across the 40 evaluated scenarios. This discrepancy is amplified under the RMSE metric, which disproportionately penalizes large residuals through squaring.  

As alluded to previously, NAP3D outperforms PnP-RANSAC in rotational error under cleaner noise conditions, achieving at most $\sim0.48^\circ$ lower error in five low-noise configurations. The persistence of a positive error difference when no blur is applied, across all intrinsics perturbations, suggests that NAP3D exhibits increased resistance to camera drift. However, as noise becomes more severe, NAP3D degrades rotationally and underperforms translationally across all tested conditions. In the worst case, this corresponds to an increase of approximately $1^\circ$ in rotational error and $14.6$ cm in translation.

Noise severity has a notable impact on relative performance. Under harsher blur and intrinsics perturbations, NAP3D tends to underperform in raw rotation and translation metrics, contrary to our initial hypothesis that PnP-RANSAC's reliance on a single 3D point set and accurate intrinsics would lead to greater instability. This effect does, however, manifest in the 3D RMSE metric: as noise increases, NAP3D improves by approximately $1.1$ cm relative to PnP-RANSAC while maintaining superior 3D alignment accuracy in all evaluated cases.

Overall, these results suggest that NAP3D provides robust and consistent 3D alignment, as evidenced by lower RMSE, while highlighting opportunities for improvement in translation and rotation estimation. While Procrustes alignment is fundamentally designed for 3D point cloud registration rather than camera pose estimation, this evaluation demonstrates that, despite this structural mismatch, a 3D-3D formulation can remain competitive with conventional 2D-3D approaches and, in some regimes, offer improved stability. These results indicate that while PnP-RANSAC may achieve lower average pose parameter error, NAP3D provides more stable 3D alignment by avoiding extreme failure modes, leading to consistently lower RMSE.

\section{Conclusion}

This paper presented a software correction system designed to reduce accumulated positional error in autonomous vehicle odometry using Neural Radiance Fields and 3D-3D keypoint alignment via Procrustes Analysis. In an initial evaluation on a custom dataset, testing in controlled environments showed that the system achieves positional RMSE below $5$ cm across multiple trials and sublocations, performing consistently better than a 2D-3D PnP baseline evaluated on the same data.

In a second evaluation on the TUM RGB-D dataset, the method was assessed under more realistic sensing noise and viewpoint variation. While a 2D-3D PnP-RANSAC baseline achieved lower raw rotational and translational parameter error in some regimes, the proposed method consistently exhibited lower 3D alignment RMSE, indicating more stable and reliable spatial alignment. Together, these results suggest that integrating 3D-3D NeRF-based corrections provides a practical and robust approach for improving localization accuracy in autonomous navigation pipelines.

While this study used staged evaluations for clarity and reproducibility, future work will focus on testing with a fully autonomous agent to evaluate the method's generalizability. The current evaluation also relies on static, unmoving scenes; future evaluations will incorporate explicit modeling for movement and dynamic environments, and further resistance to noise.

\nocite{maggio2022locnerfmontecarlolocalization}
\nocite{yenchen2021inerfinvertingneuralradiance}
\nocite{zhou2024nerfectmatchexploringnerf}
\bibliographystyle{plain}
\bibliography{refs}
\end{document}